\documentclass[11pt]{article}

\usepackage[final]{acl}

\usepackage{times}
\usepackage{latexsym}

\usepackage[T1]{fontenc}

\usepackage[utf8]{inputenc}

\usepackage{microtype}

\usepackage{inconsolata}

\usepackage{graphicx}
\usepackage{amsmath,amssymb}
\usepackage{booktabs}

\usepackage{tikz}
\usetikzlibrary{{calc}}
\usetikzlibrary{arrows.meta, bending}
\usepackage{subcaption}
\usepackage{graphicx}
\usepackage{xcolor}
\usepackage{eso-pic}

\usepackage{stfloats}
\newcommand{\inlinefig}[1]{%
  \raisebox{-0.25\height}{\includegraphics[height=3ex]{#1}}%
}
\newcommand{\lmm}{\inlinefig{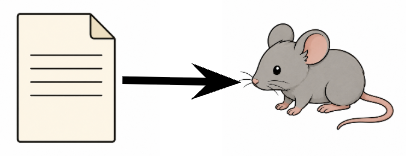}}
\newcommand{\mml}{\inlinefig{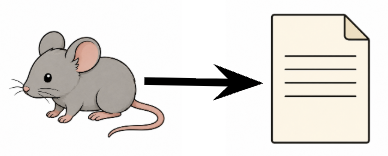}}

\newcommand{\mmm}{\inlinefig{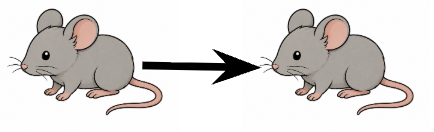}}
\newcommand{\m}{\inlinefig{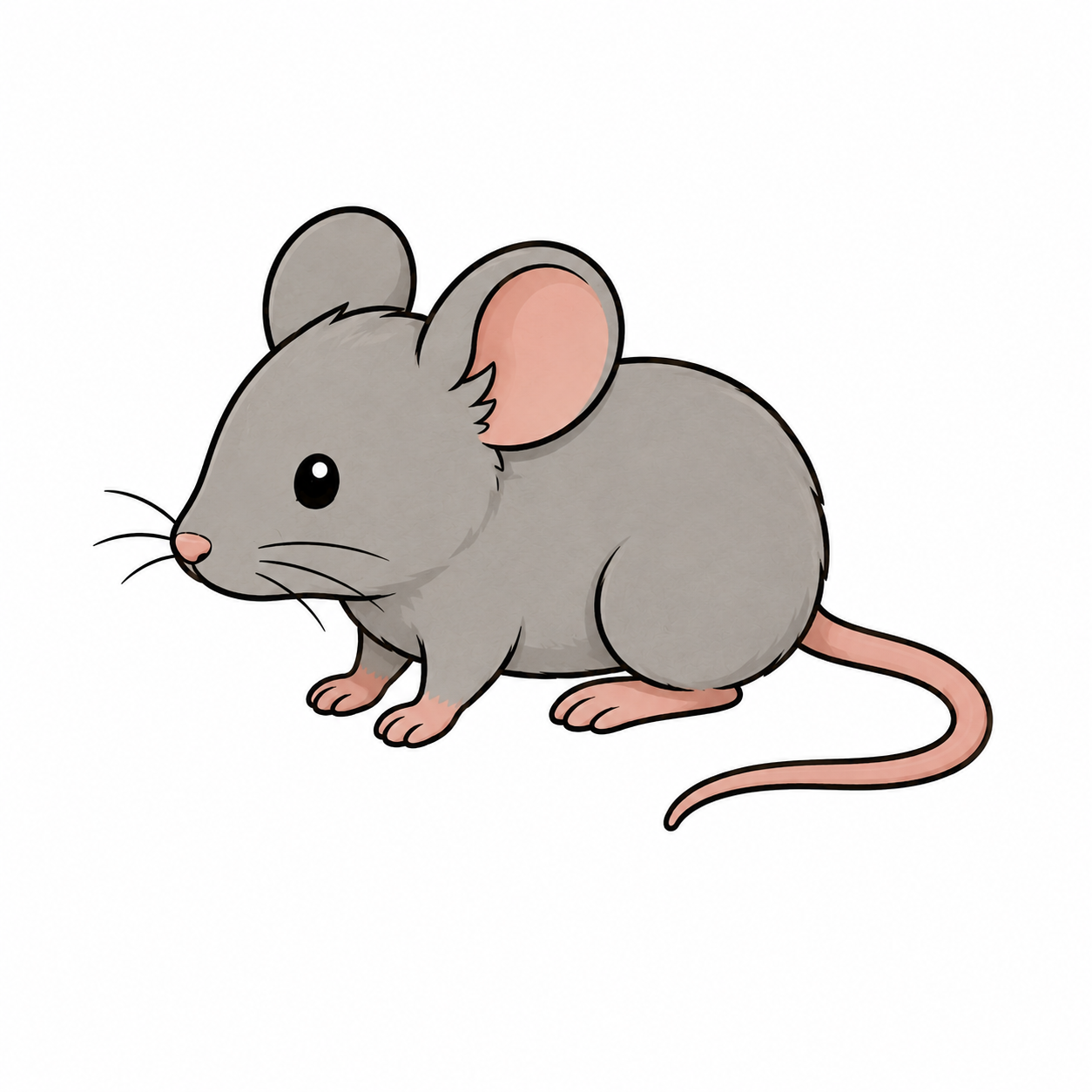}}
\newcommand{\lt}{\inlinefig{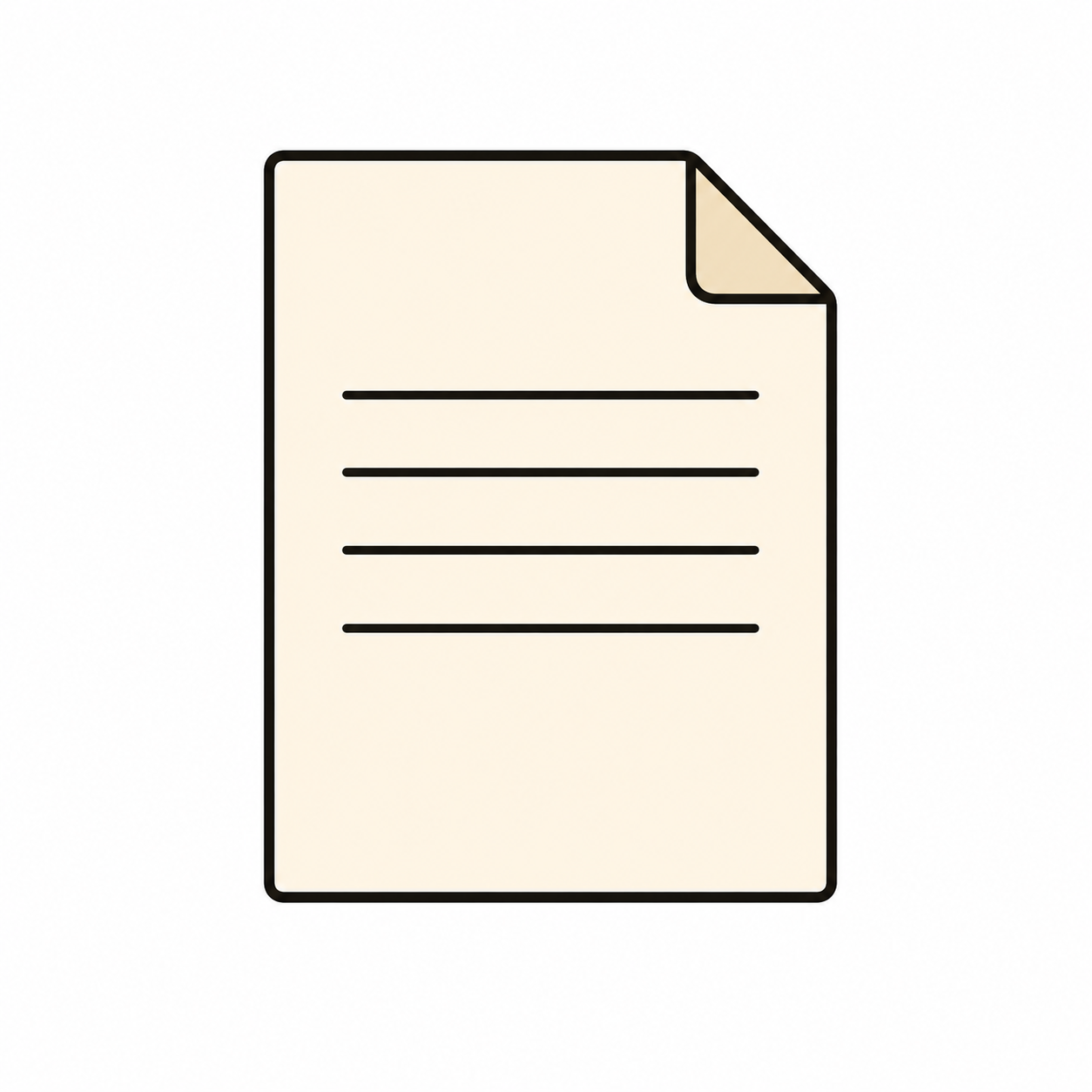}}
\newcommand{\mec}{\inlinefig{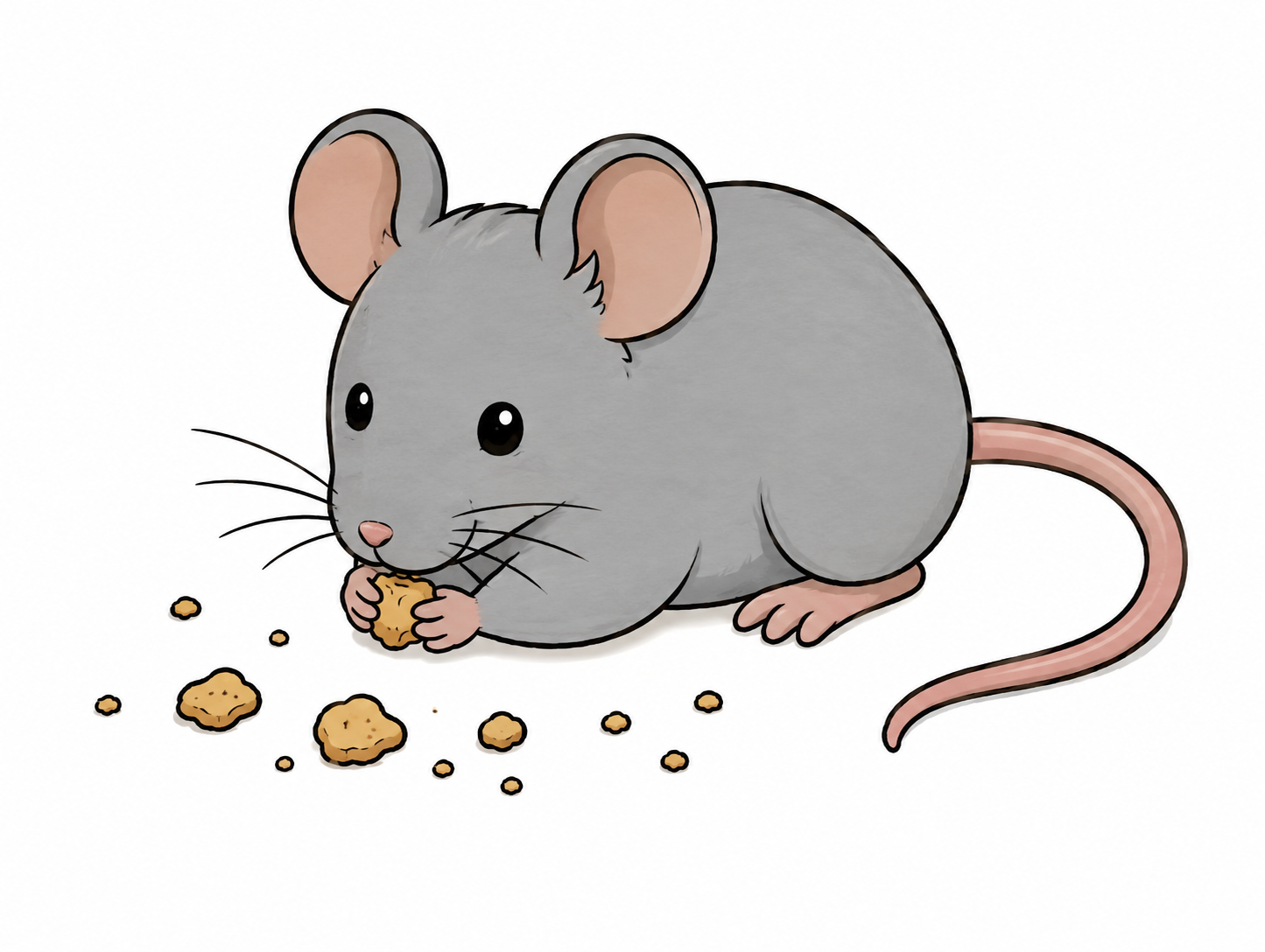}}
\newcommand{\lec}{\inlinefig{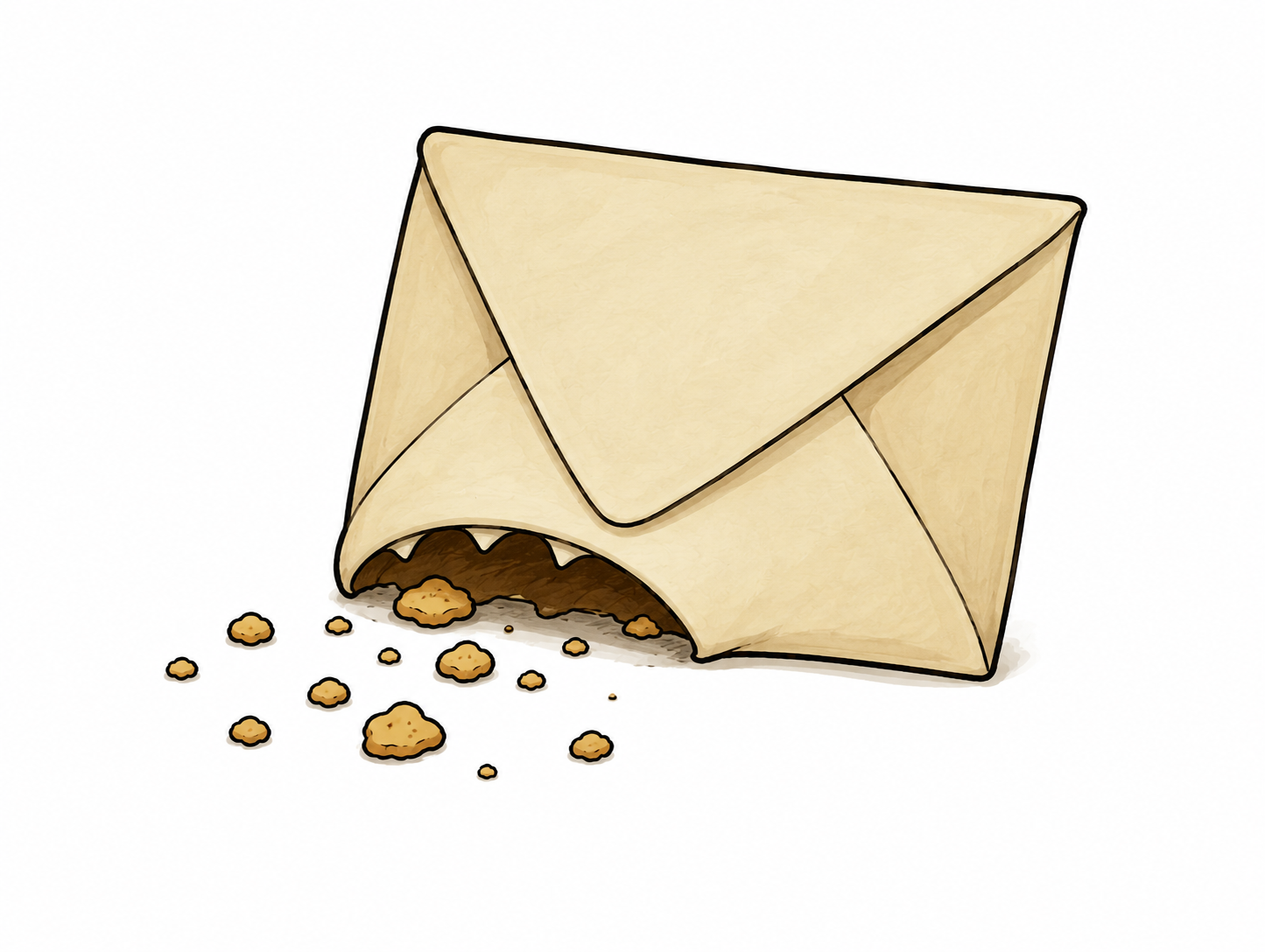}}

\usepackage[normalem]{ulem}

\title{Lexical Redefinition Reveals Context-Sensitive Instruction Following}

\author{Roksana Goworek \\
  Queen Mary University of London \\
  \texttt{r.goworek@qmul.ac.uk} \\\And
  Haim Dubossarsky \\
  Queen Mary University of London \\
  University of Cambridge \\
  \texttt{h.dubossarsky@qmul.ac.uk} \\}

\begin{document}
\pagestyle{plain}
\setcounter{page}{1}

\AddToShipoutPicture*{
  \setlength{\unitlength}{1mm}
  \put(0,280){\parbox[t]{\paperwidth}{\centering
    \footnotesize
    \emph{Accepted to the 3rd Workshop on Language Understanding
    in the Human-Machine Era (LUHME 2026)}\\
    October 28, 2026 $\cdot$ Co-located with EMNLP 2026
    $\cdot$ Budapest, Hungary
  }}
}

\maketitle
\thispagestyle{plain}
\begin{abstract}

How do language models reconcile conflicting contextual evidence within a prompt? We study this question using lexical redefinition, where an explicit instruction specifies how a word should be interpreted and a subsequent sentence provides additional evidence through its usage. Across 870 word pairs, we find that lexical redefinitions are not applied as fixed rules. Instead, the behaviour depends systematically on the relationship between the explicit instruction and the intervening sentence: the contextualised usage can preserve, weaken, or completely reverse the instruction. These results suggest that the interpretation of an explicit instruction remains sensitive to subsequent contextual evidence. Lexical redefinition therefore provides a controlled behavioural paradigm for studying how language models integrate multiple sources of information within a prompt.

\end{abstract}

\section{Introduction}
\label{sec:intro}

Prompted language models receive multiple sources of contextual information. Some are explicit, such as instructions that specify how a task should be performed \citep{mishra-etal-2022-cross}. Others are implicit, arising from examples, demonstrations, or ordinary language use elsewhere in the prompt \citep{brown2020language, min-etal-2022-rethinking}. These sources often agree, but they can also conflict in subtle ways that change how an explicit instruction should be interpreted \citep{camassa2026instruction, xu2024knowledge}. Understanding how language models reconcile such competing contextual evidence is central to understanding instruction following and in-context reasoning.

Humans likewise rarely interpret instructions in isolation. After hearing "Put the kettle on" a subsequent remark such as "I have some great biscuits" or "Do we have some orange juice left?" is naturally understood as relevant to the earlier instruction, even though it does not explicitly refer to it. In the first case, the remark explains why the instruction was given; in the second, it changes what the listener understands the speaker intended, without explicitly revising the original instruction.

Whether language models perform a similar kind of contextual reconciliation is much less clear. Existing work shows models do not treat all contextual information equally. They may disregard in-context demonstrations in favour of pretraining priors \citep{min-etal-2022-rethinking}, rely on the semantic properties of labels rather than newly specified mappings \citep{wei2023larger}, or trust contextual information selectively depending on whether it agrees with stored knowledge \citep{xu2024knowledge}. However, these studies mainly ask if models use particular sources of contextual information. Much less is known about how models reconcile multiple sources when these point toward different interpretations.

We use a lexical redefinition paradigm as a controlled experimental probe. A redefinition — "the word 'mouse' should be interpreted as 'letter'" (illustrated as \mml{}) — asks the model to override a pretrained association on command. We then insert a single intervening sentence that either demonstrates the instructed remapping of meaning, uses the redefined word in its original ordinary sense, or demonstrates the opposite remapping. This creates a controlled competition between two sources of contextual evidence: an explicit instruction that states how a word should be interpreted, and a demonstrated usage that shows how it is interpreted. Because these two sources can be manipulated independently while everything else is fixed, lexical redefinition provides a minimal setting for studying how models reconcile competing contextual evidence.

We study the effect of lexical redefinition on Qwen2.5-14B-Instruct across 870 word pairs, and show that the effect depends on the relationship between the explicit instruction and subsequent contextual evidence. The same instruction produces different behaviour depending on how it is related to the intervening sentence. 

Our findings suggest that explicit instructions are reconciled with implicit contextual evidence, rather than processed independently. We report that a single sentence can flip an explicit instruction when its usage points the other way, and conclude that this may indicate that the model's output tracks the \emph{relationship} between these two sources, not the instruction in isolation.

\section{Related Work}
\label{sec:related}


A closely related line of work studies conflicts between explicit instructions and patterns demonstrated in context. \citet{camassa2026instruction} show that models can shift from instruction following to pattern following after repeated contradictory demonstrations, and that this susceptibility is not well predicted by standard capability benchmarks. Our setting differs by introducing competing evidence through a single intervening sentence, varying its relation to the instruction beyond direct conflict, and focusing on how later context changes the interpretation of an instruction rather than whether the model simply complies with it.

\subsection{Conditional Use of Prompt Information}
\label{sec:related:prompt-use}

Prior work shows that models use prompt information selectively rather than learning arbitrary mappings from context. \citet{min-etal-2022-rethinking} find that randomising input--label mappings in demonstrations often has little effect, suggesting that models rely more on label semantics, input distributions, and task format to identify latent concepts acquired during pretraining \citep{xie2022explanation}. Later studies also show that pretrained semantic priors constrain the mappings models acquire in context \citep{wei2023larger,kumar2024semantic,jang2024rectifying}.

Our setting shifts the focus from whether contextual information is used to how multiple relevant cues interact. Unlike work on irrelevant or distracting context \citep{shi2023large,yoran2024making}, the intervening sentence is always pertinent to the redefinition but varies in its relation to it. We therefore frame prompt interpretation as an interaction between an explicit rule and a single demonstrated usage, showing that model behaviour depends on the relationship between these two in-context sources. This also differs from knowledge-conflict settings \citep{xu2024knowledge}: both competing signals are contextual, while parametric knowledge provides only the default mapping that the redefinition attempts to override.

\subsection{Semantic Plausibility and Anomaly}
\label{sec:related:anomaly}
Prior work shows that language models are sensitive to semantic plausibility, but less reliably capture finer-grained contextual expectations \citep{ettinger2020bert,kauf2023event}. Models can also interpret anomalous but coherent sentences when an appropriate context is available \citep{olsen2025finding}. Our stimuli involve a different form of anomaly: rather than violating world knowledge, they place one word in a context where another would ordinarily occur, creating a distributionally unexpected but interpretable usage.

We use this controlled anomaly as contextual evidence about a lexical redefinition. Its effect depends on its relation to the explicit instruction: ordinary usage has limited influence, whereas an anomalous usage that supports an alternative mapping can substantially alter or override the instructed interpretation.

\section{Methods}
\label{sec:methods}

\subsection{Stimuli}
We use 30 words, evenly selected from three part-of-speech groups (10 nouns, 10 verbs, 10 adjectives) and chosen to reflect various lexical properties such as concreteness, animacy, and polysemy. We construct all $30\times29=870$ ordered pairs, since $X\!\rightarrow\!Y$ and $Y\!\rightarrow\!X$ are distinct redefinitions. The full word list is given in Appendix~\ref{tab:wordlist}. 
For each word, Gemini-2.5-Pro generates 10 natural sentences of 8--15 words in which the word occurs naturally.\footnote{Shorter examples are used for illustration (e.g., \textit{``The mouse eats crumbs''}); the actual sentences are longer.}



\paragraph{Ordinary and Anomalous Sentences}
These sentences generated by Gemini serve as \emph{ordinary} intervening sentences and as templates for \emph{anomalous} variants. For a pair such as (\textit{mouse}, \textit{letter}), the ordinary sentence contains the first word in its natural context (\textit{``The mouse eats crumbs''} \mec), while the anomalous variant replaces it with the second word (\textit{``The letter eats crumbs''} \lec). Thus, the surrounding context remains fixed and only the target word changes. Sentences are presented without an explicit link to the redefinition.

\subsection{Prompts}
Depending on the condition, each sentence is preceded by a redefinition prompt of the form:
\begin{list}{}{\leftmargin=0.4cm \rightmargin=0.4cm}
\item
\small\itshape
``In the following sentences, the word `\,$\langle$A$\rangle$\,' should be
interpreted as if it were the word `\,$\langle$B$\rangle$\,'. Please answer
questions based on this interpretation.''
\end{list}


We tested four alternative prompt formulations on 56 word pairs to ensure that the results do not depend on this specific prompt. The full prompts can be found in Table~\ref{tab:wordings} of the Appendix.


\subsection{Sentence Completion Task}
\label{sec:methods_completion}
Depending on the condition, (\mml{}, \lmm{},or "$\varnothing$") each prompt or intervening sentence is followed by an incomplete sentence-completion query whose natural continuation is one of the two words in the pair (e.g.\ \textit{``The cat chased the tiny gray
\rule{1.2em}{0.4pt}''} for \textit{mouse}). We record the probability assigned to each word as the next token. Queries are generated using Claude Opus~4.7 (one query per word), and reviewed manually to ensure the intended word is the natural completion and the sentence reads naturally.



\subsection{Experimental Design}
The design is fully counterbalanced. For every word pair, we include both redefinition directions and construct ordinary and anomalous sentences for both words. Thus, alongside an ordinary \emph{mouse} sentence and an anomalous \emph{letter} sentence, the dataset also contains an ordinary \emph{letter} sentence and an anomalous \emph{mouse} sentence. The same applies to all 30 words. Sentence-completion queries are counterbalanced in the same way. We use only one orientation of the \emph{mouse}--\emph{letter} example in the main text for readability, but all analyses include the corresponding reverse cases and group results by experimental role rather than lexical identity.

\subsection{Model}
We evaluate Qwen2.5-14B-Instruct. Exact Hugging Face checkpoint identifiers and details of the probability extraction procedure, including aggregation across whitespace and capitalisation variants, are provided in Appendix~\ref{app:implementation}.



\subsection{Analyses}
\label{sec:methods_analyses}
All analyses below are based on item analyses, where we compare probabilities of the same pairs across the different conditions. This allows us to directly, and fairly, measure the effects of the experimental conditions, as variance associated with different base probabilities across the 870 pairs is cancelled out.

\paragraph{Token Probabilities}
Each experimental condition is defined by a redefinition prompt, an intervening
sentence (or none), and a query. We evaluate each condition on ten sentence
instances and report the mean probability

\begin{equation}
P(t)=\frac{1}{10}\sum_{i=1}^{10}p^{(i)}(t),
\end{equation}

where $p^{(i)}(t)$ is the probability assigned to token $t$ for sentence instance
$i$. Throughout, $P(t)$ denotes this condition-level average.

\paragraph{Sentence Effect}
\label{sec:methods_effects}

\begin{figure}[t]
  \centering
  \includegraphics[width=\linewidth]{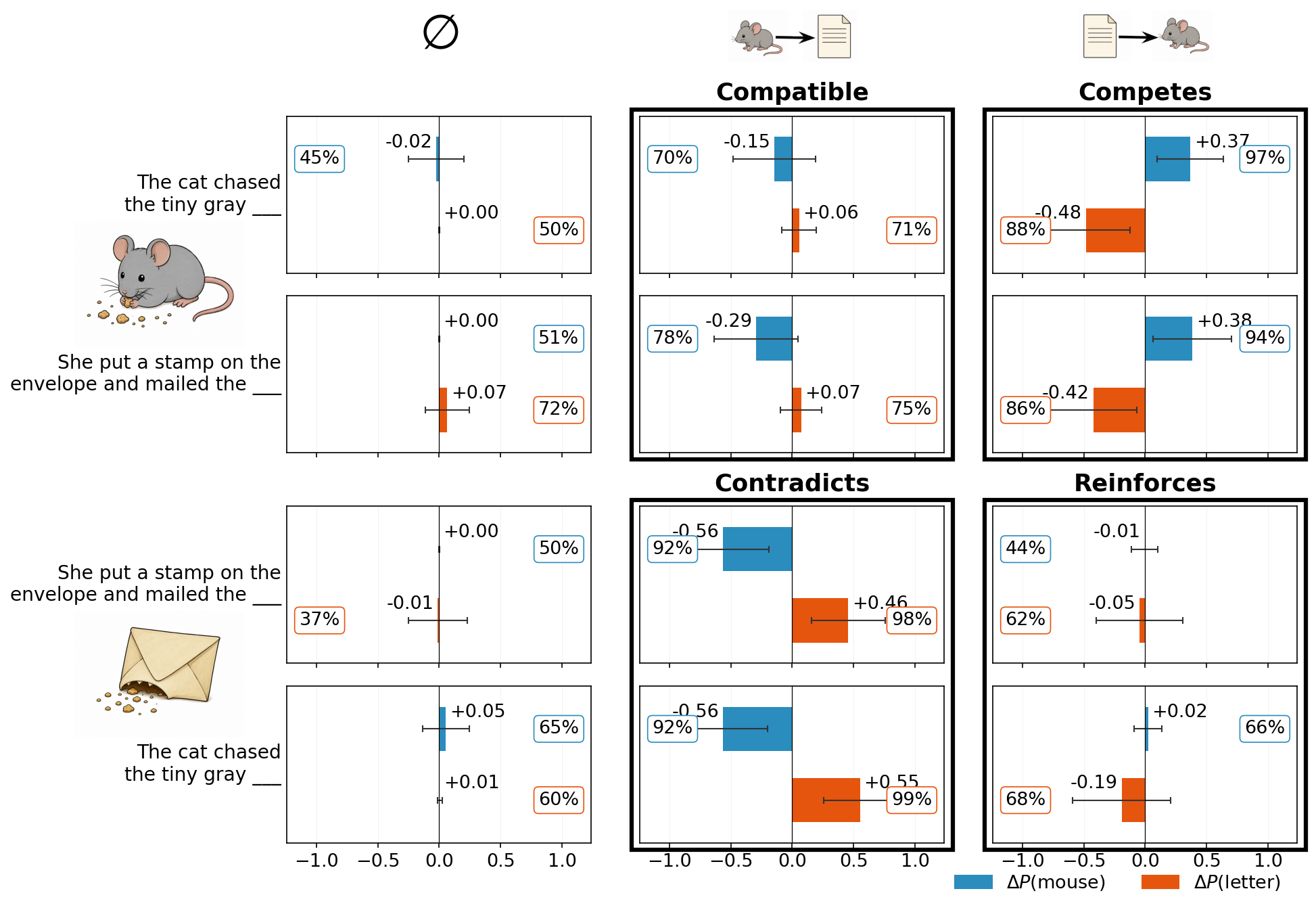}
  \caption{Effect of the intervening sentence on completion probabilities (\(\Delta P\), Eq.~\ref{eq:dP}). Blue/orange bars show \(\Delta P\) for \emph{mouse}/\emph{letter}. Percentages inside boxes indicate sign agreement with the mean across word pairs (Eq.~\ref{eq:sign}).}
  \label{fig:main}
\end{figure}

To isolate the effect of the intervening sentence, we compare each condition with an otherwise identical prompt containing no intervening sentence (only the redefinition prompt and completion query).
We define the sentence effect as the difference in completion probability between the sentence and matched no-sentence conditions, holding the redefinition and query fixed:
\begin{equation}
  \Delta P(t) \;=\; P_{\text{sent}}(t) - P_{\text{no-sent}}(t).
  \label{eq:dP}
\end{equation}
This is the quantity reported in Figure~\ref{fig:main}.



\paragraph{Sign Agreement}
\label{sec:methods_signagreement}
As a descriptive measure of directional consistency across word pairs, we
report sign agreement. For each condition $c$ and candidate token $t$, we
first compute the aggregate mean sentence effect
$\overline{\Delta P}_c(t)$ across the 870 ordered word pairs. We then
measure the proportion of pairs whose individual sentence effect has the
same direction as this aggregate mean. Pairs with exactly zero effect
contribute $0.5$, reflecting no evidence for either direction. Formally,

\begin{equation}
A(t)=\frac{1}{870}\sum_{j=1}^{870}s_j(t),
\label{eq:sign}
\end{equation}
where $j$ indexes word pairs and
\[
s_j(t)=
\begin{cases}
0.5, & \text{if } \Delta P_j(t)=0,\\
1, & \text{if } \operatorname{sign}(\Delta P_j(t))
=\operatorname{sign}(\overline{\Delta P}(t)),\\
0, & \text{otherwise.}
\end{cases}
\]

Here, $\Delta P_j(t)$ is the sentence effect for word pair $j$ on the probabilities of token $t$, and $\overline{\Delta P}(t)$ is the mean sentence effect across all pairs. Values near $50\%$ indicate little consistency in the direction of the effect, whereas values approaching $100\%$ indicate that most pairs change in the direction of the mean. The percentages are shown beside the corresponding bars in Figure~\ref{fig:main}.








\paragraph{Aggregation} Probabilities are first averaged across the ten sentence instances belonging to a word pair, yielding one score per pair. Means and standard deviations are then computed across the 870 word pairs. This reduces variation due to the particular sentence instances, and standard deviations reflect differences between word pairs.


\section{Results}
\label{sec:results}

We report main results in the text, and present the prompt-variation results in Appendix \ref{sec:phrasing-result}. 
We report effect magnitudes and directional agreement across word pairs descriptively. Because lexical items, sentences, and completion queries recur across ordered pairs, the pair-level observations are not fully independent (see Limitations section).

\subsection{Redefinition Effect} 
\label{sec:Results_redef_effect}

Before introducing intervening sentences, we verify that the redefinition instruction alone changes the model's completions as expected.

\begin{figure}[b]
    \centering
    \includegraphics[width=\linewidth]{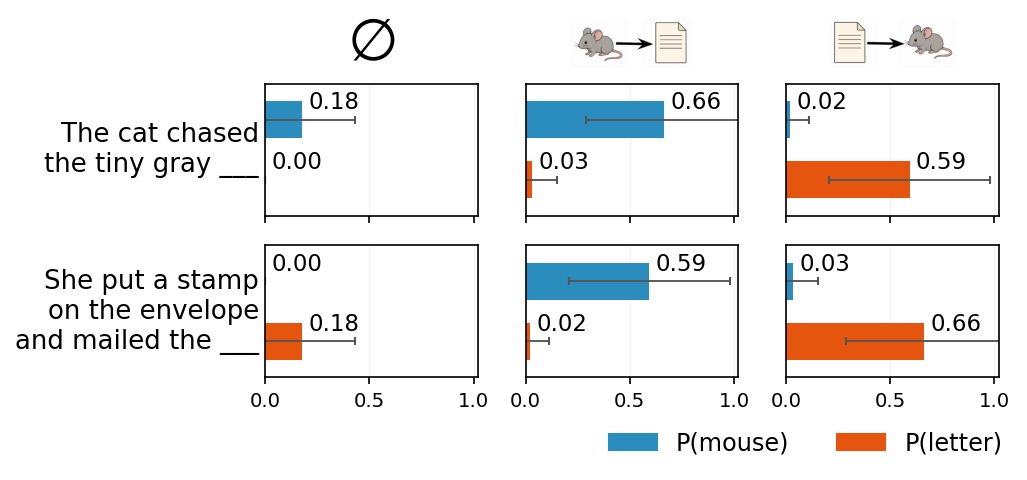}
    \caption{Token probabilities for sentence-completion queries under redefinition prompts and the original prompt without redefinition ("$\varnothing$"). Sentences and icons are illustrative, results are aggregated over 870 word pairs. Error bars show SD.}
    \label{fig:nosentence}
\end{figure}

Figure~\ref{fig:nosentence} shows the averaged completion probabilities before and after the redefinitions were applied. Instructing the model with the redefinition \lmm{} ("letter means mouse") makes it generate responses consistent with the instructed mapping. For the query \textit{"The cat chased the tiny gray \underline{\hspace{.5cm}}}" , where mouse is the natural completion, the redefinition increased P(\lt) from 0\% to 59\%, while the probability of what was the original correct completion (\m) dropped from 18\% to 2\%. Similarly, the redefinition \mml\ increased P(\m) from 0\% to 59\% for the query \textit{"She put a stamp on the envelope and mailed the \underline{\hspace{.5cm}}}", where letter is the natural completion, while dropping the original correct completion P(\lt) from 18\% to 2\%. By completely flipping the probabilities between the correct sentence completion in the original case and the redefinition, our results show that the model applied the instructions correctly.

Furthermore, the same instructions do not ``flip'' the probabilities in cases where their redefinitions do not make sense. \lmm{} only works for queries with mouse sentences and not letter sentences, and conversely, \mml{} only works for letter sentences and not mouse sentences. Instead, in these cases we observe an increase in P(\m{}) and P(\lt{}) relative to their baseline. We interpret this as a priming effect \citep{kassner-schutze-2020-negated}, where a token that appears in a previous context can affect the generated probabilities. Because subsequent analyses measure sentence effects relative to matched no-sentence baselines (Eq.~\ref{eq:dP}), this priming effect is controlled for by design.

\subsection{Intervening Sentence Effect}

We next analyse the effect of the intervening sentence on the next token probabilities for the sentence-completion query. We stress that the sentence is simply placed there, without any explicit instructions. A priori, one may think that the intervening sentence would imply a meaning associated with the particular usage of the word in the sentence. An anomalous sentence of the form \lec{} implies that \textit{letter} means \textit{mouse} (a different meaning) by showing an example of letter in a sentence where mouse would normally occur. Conversely, an ordinary sentence \mec{} implies that \textit{mouse} means \textit{mouse} (its current meaning) by showing it in its natural context. If the model follows the \textit{meaning} that is implied by the sentence (but never explicitly stated like in the redefinition), we would expect probabilities assigned to tokens to change, compared with the matched condition containing no intervening sentence. For example, since the \lec{} sentence implies that \lmm{}, then if the model follows this meaning implicature, we would expect it to complete the sentence "The cat chased the tiny gray \_\_\_" with \textit{letter}, in line with the implied meaning, and similar to the redefinition effect of the equivalent mapping observed in \S~\ref{sec:Results_redef_effect}.

Contrary to that expectation, we observe little evidence of the implied mapping in the $\varnothing$ column of Figure~\ref{fig:main}, which shows the effect of the sentence on completion without any redefinition instruction. In fact, presenting the \lec{} sentence before asking the model to complete the sentence "The cat chased the tiny gray \_\_\_" results in a 5-percentage-point-increase in P(\m), relative to the no-sentence baseline (shown in Figure~\ref{fig:nosentence}), with little corresponding change in P(\lt). The pattern therefore does not support adoption of the mapping implied by the sentence: the ordinary completion increases more than the substituted word.

We therefore find no evidence that presenting a sentence by itself has a meaningful impact on the model's next token probabilities. It seems that the model does not take on the implied redefinition of the sentence and instead outputs nearly the same token probabilities as it did without any sentence.



\subsection{Sentence x Redefinition Interactions}

We next examine the interaction between the redefinition and the intervening sentence. Although the prompt does not explicitly link them, the sentence substantially alters predictions after a redefinition, despite having almost no effect on its own.

The direction and magnitude of this interaction depend on how the sentence relates to the instructed mapping. Combining the two redefinition directions with the two sentence types yields four distinct sentence--instruction relations: reinforcing, contradicting, competing, and compatible. These relations are shown in Figure~\ref{fig:main} and summarised in Table~\ref{tab:relationships} and constitute the paper's main finding.

\begin{table*}[b]
\centering
\small
\begin{tabular}{@{}lp{5.6cm}p{5.6cm}@{}}
\toprule
Sentence
& \includegraphics[height=1.2em]{images//mouse_means_letter.jpeg}
& \includegraphics[height=1.2em]{images//letter_means_mouse.jpeg} \\
\midrule
\begin{tabular}[t]{@{}c@{}}
\raisebox{-\height}{\includegraphics[height=3em]{images//mouse_eats_crumbs.jpeg}} \\
\textit{``The mouse eats crumbs''} \\
$\therefore implies$ \mmm
\end{tabular}
& \textbf{Compatible} \newline uses \m in its ordinary sense,
but the\newline redefinition affects only \lt contexts \newline $\rightarrow$ redefinition unaffected or slightly \newline weakened
& \textbf{Competes} \newline uses \m in its ordinary sense,
which was \newline redefined to be denoted by \lt \newline $\rightarrow$ \mml adopted instead \\
\addlinespace
\begin{tabular}[t]{@{}c@{}}
\raisebox{-\height}{\includegraphics[height=3em]{images//letter_eats_crumbs.jpeg}} \\
\textit{``The letter eats crumbs''} \\
$\therefore implies$ \lmm
\end{tabular}
& \textbf{Contradicts} \newline shows an opposite mapping to the one in the instruction  \newline $\rightarrow$ \lmm adopted instead
& \textbf{Reinforces} \newline shows an example of the redefinition \newline $\rightarrow$ redefinition effect is unchanged \\
\bottomrule
\end{tabular}
\caption{Analysis of the four sentence--redefinition combinations.
Each cell names the relationship between the sentence and the redefinition instruction, and the observed effect on the completions.}
\label{tab:relationships}
\end{table*}

\subsubsection{Anomalous Intervening Sentences}

\noindent An anomalous intervening sentence places one word in a semantic context where the other would ordinarily occur, thereby implying a lexical mapping without stating it explicitly. For example, \lec{} uses \emph{letter} in a mouse context and therefore implies \lmm{}. This implied mapping provides a possible explanation for the different effects observed in the reinforcing and contradicting conditions.

\paragraph{Reinforcing}
In this condition, the intervening sentence produces almost no change to the completion probabilities relative to the no-sentence redefinition condition. For example, under \lmm{} followed by \lec{}, the model applies the redefinition with approximately the same strength as when no intervening sentence is present, and the token probabilities remain largely unchanged. A possible explanation is that the anomalous sentence implies the same mapping as the explicit instruction. The redefinition states \lmm{}, and \lec{} also suggests that \emph{letter} denotes the mouse concept. Because the explicit and implied mappings align, the sentence introduces no competing evidence and therefore does not alter the model's response.

\paragraph{Contradicting}
In this condition, the intervening sentence produces a reversal in the model's predictions. When \mml{} is followed by \lec{}, P(\m) decreases sharply while P(\lt) increases substantially (see the bottom-left quadrant of Figure~\ref{fig:main}). The probabilities do not merely shift; they flip, closely resembling the pattern observed under the opposite \lmm{} redefinition in the no-sentence condition (Figure~\ref{fig:nosentence}). A possible explanation is that the model is acting upon the reverse of the instructed mapping implied by the anomalous sentence. The instruction states \mml{}, whereas \lec{} suggests \lmm{}. Because these mappings assign opposite word--concept relations, they directly contradict one another. The observed probability reversal could thus reflect the model resolving this conflict in favour of the mapping implied by the intervening sentence.

\subsubsection{Ordinary Intervening Sentences}

Ordinary intervening sentences differ from anomalous ones because they do not imply a new lexical mapping, but preserve an existing word--concept relation. Their effect on completion probabilities depends on whether this relation concerns the concept targeted by the redefinition or the word being redefined. An ordinary sentence either shows that the concept targeted by the redefinition is still expressed by its original word, or that the redefined word still retains its original meaning. These two cases correspond to the competing and compatible conditions, respectively.

\paragraph{Competing}
A probability reversal similar to that in the contradicting condition, though arising for different reasons, occurs here. Under \lmm{}, inserting the ordinary sentence \mec{} flips the preferred completion probabilities from \emph{letter} back to \emph{mouse}. The intervening sentence presents the \m{} concept as being expressed by its original word, \emph{mouse}, whereas the redefinition instructs the model to express that concept using \emph{letter}. It therefore provides direct evidence against the instructed mapping, but without directly showing the opposite mapping instead.

\paragraph{Compatible}
A different pattern is observed for \mml{} followed by the ordinary sentence \mec{}. Here, the intervening sentence implies that \emph{mouse} retains its ordinary meaning, but provides no evidence about which word should denote the \lt{} concept. Nevertheless, the results show that the preference for \emph{mouse} decreases significantly, while the probability of \emph{letter} does not change correspondingly and the token probabilities do not flip. We interpret this as a dilution effect. The redefinition initially makes \emph{mouse} a strong candidate for the letter concept, but the ordinary sentence provides evidence that \emph{mouse} also retains its original meaning. This makes its association with the letter concept less specific, reducing P(\m).

\section{Discussion}
\label{sec:discussion}

Lexical redefinition provides a controlled setting for studying how language models reconcile multiple sources of contextual evidence. Across our experiments, the model did not treat an explicit redefinition as either a fixed rule or an instruction to be ignored. Instead, it integrated the instruction with evidence supplied by a later intervening sentence, and its predictions depended systematically on the relation between the two sources.

This pattern cannot be explained by either source alone. Redefinitions strongly altered completion probabilities in the expected direction, whereas intervening sentences had little effect without an explicit instruction. Yet when combined, a single sentence could preserve, attenuate, or reverse the redefinition effect. The interaction was also robust across alternative prompt formulations, suggesting that it reflects the relation between instruction and usage rather than a particular wording.

We propose four categories---reinforcing, contradicting, competing, and compatible---as an explanatory framework for these distinct behavioural patterns (Table~\ref{tab:relationships}). A reinforcing sentence leaves the instruction-induced preference largely unchanged. Both contradicting sentences, which support the reverse mapping, and competing sentences, which preserve the target concept's conventional name, reverse that preference, although with different magnitudes. Compatible sentences instead produce a weaker attenuation without changing the preferred completion.

The same ordinary and anomalous sentences can have sharply different effects depending on the redefinition they follow, and these relations are asymmetric. Although the model was not instructed to compare the two sources, its predictions varied systematically with the specific logical relation between the mappings supported by the instruction and the intervening sentence.

This pattern is consistent with later context revising the behavioural consequences of an earlier instruction, suggesting that prompt interpretation remains open to revision rather than being fixed at the point of instruction. However, the fit between the results and our four-category framework does not establish that models explicitly represent or compute these categories. Rather, the framework provides a compact logical account of how different forms of agreement and disagreement between explicit rules and contextual evidence produce distinct behavioural outcomes.

\paragraph{Unlabelled Context Influences Instruction Use}
The intervening sentence is never introduced as evidence about the redefinition. Nevertheless, it systematically modulates the effect of the earlier instruction. This invites comparison with relevance-theoretic accounts of human communication, according to which communicative inputs create expectations of relevance and are interpreted against available contextual assumptions \citep{sperber-wilson-1995-relevance,
wilson-sperber-2004-relevance}. We intend this only as a behavioural analogy: the present results do not show that language models implement the cognitive mechanisms proposed by relevance theory, leaving this as an open question.


\paragraph{Implications}
An explicit instruction is therefore not necessarily the strongest signal in a prompt. This is in line with \citet{camassa2026instruction}, who show that sustained in-context demonstrations can override an explicit instruction; here, the same reversal is produced by a single intervening sentence. 
The result is also compatible with accounts in which in-context behaviour depends on properties such as the demonstrated label space, sequence format, or inference over a latent task concept \citep{min-etal-2022-rethinking,xie2022explanation}.
More generally, prompt-injection research shows that models can treat contextual information as behaviour-changing instructions even when that data is not intended to have instructional authority \citep{greshake2023not}. Recent work on in-context representation hijacking shows that repeated lexical substitutions can shift how a token is represented and interpreted in adversarial prompts \citep{yona-etal-2026-context}. Our results identify a related sensitivity in a controlled, non-adversarial setting. Whether it extends to realistic security scenarios remains a topic for future work.

\paragraph{Lexical Semantic Change Research} Computational approaches to lexical semantic change are grounded in the assumption that differences in usage provide evidence for differences in meaning \citep{kutuzov-etal-2018-diachronic, goworek-dubossarsky-2026-rethinking}. Recent work uses computational models, including LLMs, to make usage-level semantic judgements \citep{schlechtweg2025comedi,de-sa-etal-2026-semantic,spaziani-etal-2026-elections}. Although our experiments do not directly evaluate LSC annotation, they point to a consideration for this emerging methodology: an LLM’s interpretation of a usage may depend not only on the usage itself, but also on other information in the prompt. As LLM-based annotation becomes increasingly capable and widely adopted, improvements in apparent annotation accuracy may be worth considering alongside possible context-dependent artefacts. Future work could test biases in LLM-based semantic-change judgements due to preceding context, prompt formulation, and demonstrations.

\paragraph{Open questions} 

The observed responses are consistent with the model combining explicit instructions with later contextual evidence, but they do not distinguish among several possible explanations. The model might infer a lexical mapping, condition on demonstrated usage or rely on broader pragmatic expectations. For example, contextual entrainment increases the probability of tokens that appeared earlier in the prompt \citep{niu2025entrainment}. Determining whether the present results rely on this process, a semantic mapping process, or an interaction between them requires mechanistic analysis.

Existing work shows that in-context demonstrations can induce task- or function-level internal representations that causally affect model behaviour \citep{hendel-etal-2023-context,todd2024function}. Similar analyses could test whether lexical redefinitions and intervening usages produce separable representations, and how those representations interact.

More broadly, it remains unclear how specific the phenomenon is to lexical redefinition. The present experiments isolate competition between an explicit instruction and an implicitly demonstrated usage, but similar tensions arise in many prompting settings. Anomalous-but-interpretable sentences may make this competition especially visible; related work shows that language models can often construct supporting contexts that render anomalous sentences interpretable \citep{olsen2025finding}. Future work should test how far the effect generalises across instruction types, languages, word classes, invented vocabularies, and model scales. Lexical redefinition may therefore provide a useful controlled instance of a broader form of interaction between explicit instructions and contextual evidence.

\section{Conclusion}
\label{sec:conclusion}






In this paper, we use lexical redefinition as a controlled probe of how competing sources of contextual evidence are reconciled. An explicit redefinition and an intervening sentence each provide evidence about how a word should be interpreted, and varying them independently shows that these two sources are integrated rather than the redefinition being followed literally. A sentence that has no effect on its own can preserve, weaken, or reverse the instruction depending on the relationship between them. We capture these patterns with a four-category explanatory framework—reinforcing, contradicting, competing, and compatible—even though the model was never instructed to distinguish these relations. Its outputs nevertheless appear to track this logical structure. This suggests that models may treat redefinitions not as fixed rules to be executed but as evidence that remains open to revision. The paradigm reduces a broad question about in-context reasoning to a minimal setting in which the competing signals can be varied and measured directly.

\section*{Limitations}
\label{sec:limitations}

The lexical redefinition paradigm isolates a simple form of conflict
between an explicit instruction and a single piece of contextual evidence.
This controlled design enables systematic comparisons, but it remains
unclear how directly the results extend to natural prompting settings, in
which instructions, demonstrations, and contextual cues are more varied.
The experiments also use a small hand-selected English vocabulary and
short sentences. The
prompt-wording variation analysis is restricted to a 56-pair subset. The present
study should therefore be understood as a controlled behavioural
demonstration rather than evidence of broad generality.

\paragraph{Model Scope}
We evaluate only a single model, Qwen2.5-14B-Instruct. The extent to which the observed effects generalise to other model families, parameter scales, training regimes, or multilingual systems therefore remains unclear. In particular, larger models or models with different instruction-tuning procedures may resolve semantic conflict differently. We also measure direct next-token predictions without eliciting intermediate reasoning, so the reported context dependence may change under alternative inference procedures.

\paragraph{Construct of the Redefinition}
Our instruction targets semantic interpretation
(\emph{``A should be interpreted as B''}), whereas the sentence-completion
task measures lexical production. Completion probabilities therefore provide
a behavioural proxy for whether the model applies the redefinition, but they
do not uniquely distinguish a change in semantic interpretation from a
change in production preference or response strategy. This does not affect
the observed context sensitivity of the completion probabilities, but it limits the conclusions that can be drawn about changes to the model's internal representation of word meaning. 
Moreover, the
task measures next-token probabilities for two prespecified candidates
rather than the model's full output distribution or an independent measure
of lexical interpretation. Some variability, particularly in non-matching
queries, may therefore reflect a mismatch between the instructed construct
and the elicited behaviour. Denotational prompts or tasks that separate
interpretation from production could help disambiguate these possibilities.

\paragraph{Stimuli}
We construct 870 ordered pairs from a hand-selected set of 30 words
spanning nouns, verbs, and adjectives. This provides systematic
combinatorial coverage but limited lexical diversity, and the pairs are not
870 wholly independent lexical items because the same words, sentences,
and completion queries recur across ordered pairs. Because all words are
crossed, many pairs also span parts of speech, so some substitutions
introduce grammatical-category mismatches in addition to semantic anomaly.

The intervening sentences are generated with Gemini~2.5~Pro and reviewed
manually, but generation at scale limits experimental control. We use one
completion query per word, and the intended completion is not guaranteed
to receive the highest baseline probability for every item. Some
substitutions intended to be anomalous may also remain locally plausible
(e.g.\ substituting \emph{hot} for \emph{cold}). Measuring sentence effects
relative to matched no-sentence baselines mitigates but does not eliminate
these sources of variation.


\paragraph{Behavioural Focus}
The study characterises output behaviour and does not identify the internal
representations or computations responsible for the observed probability
shifts. In particular, the results do not distinguish lexical
reinterpretation from contextual entrainment, production-level priming, or
other forms of contextual conditioning. Mechanistic analysis would be
required to determine how the instruction and intervening sentence are
represented and how they interact during inference.

\section*{Acknowledgements}
This work has in part been funded by the research program Change is Key! supported by Riksbankens Jubileumsfond (under reference number M21-0021).


\appendix

\section{Stimuli}
\label{app:stimuli}
The 30-word set is hand-selected to balance parts of speech (10 nouns, 10 verbs,
10 adjectives; Appendix~\ref{tab:wordlist}). The
8-word subset used for the prompt-variant experiments is
\{chair, dog, flower, bank, river, cook, stone, cloud\}, giving $8\times7=56$
ordered pairs.

\begin{table}[ht]
  \centering
  \small
  \begin{tabular}{lll}
    \toprule
    Nouns & Verbs & Adjectives \\
    \midrule
    truth  & run      & hot   \\
    chair  & know     & cold  \\
    dog    & break    & big   \\
    flower & give     & small \\
    bank   & think    & good  \\
    river  & find     & bad   \\
    cook   & sleep    & dead  \\
    stone  & build    & happy \\
    cloud  & running  & fast  \\
    time   & knowing  & red   \\
    \bottomrule
  \end{tabular}
  \caption{The 30-word stimulus set by part of speech.}
  \label{tab:wordlist}
\end{table}


\section{Implementation details.}
\label{app:implementation}
We use the Hugging Face checkpoint \\
\texttt{Qwen/Qwen2.5-14B-Instruct}.
Completion probabilities are obtained by applying a softmax to the model
logits at the final input position. To account for tokenisation and
capitalisation differences, for each candidate word we tokenise four
surface forms: lowercase and initial-capitalised variants, each with and
without a leading whitespace. We take the first token ID of each form and
define the candidate probability as the sum of the probabilities assigned
to the resulting distinct token IDs. This measures the probability that
the next generated token begins with one of these surface forms of the candidate
word.

\section{Prompt Variation}
\label{app:prompts}
To assess whether the main results depend on the exact phrasing of the redefinition instruction, we repeated the experiment on 56 word pairs using four paraphrased prompts with the same intended meaning. The prompt variants are listed in Appendix~\ref{tab:wordings}, and the corresponding results are shown in Appendix~\ref{tab:vardelta-scope-global}-\ref{tab:vardelta-comprehension-only}.
The qualitative pattern is consistent across all prompt variants, indicating that the interaction between the redefinition and the intervening sentence is robust to instruction wording. 



\begin{table*}[t]
\centering
\small
\begin{tabular}{@{}lp{13cm}@{}}
\toprule
Wording & Prompt (X, Y stand for the word pair) \\
\midrule
baseline & In the following sentences, the word `X' should be interpreted as if it were the word `Y'. \\
scope\_sentence\_only & For the next sentence only, `X' should be interpreted as if it were `Y'. \\
scope\_global & For the rest of this conversation, `X' should be interpreted as if it were `Y'. \\
denotational & Whenever `X' appears, read it as `Y'. \\
comprehension\_only & Treat `X' as having the meaning of `Y' when reading. Do not change how you produce these words; only how you interpret them. \\ 
\bottomrule
\end{tabular}
\caption{The baseline redefinition prompt and four alternative formulations.}
\label{tab:wordings}
\end{table*}

\section{Raw probabilities}
\label{app:rawprobs}
\begin{figure}[hp]
  \centering
  \includegraphics[width=\linewidth]{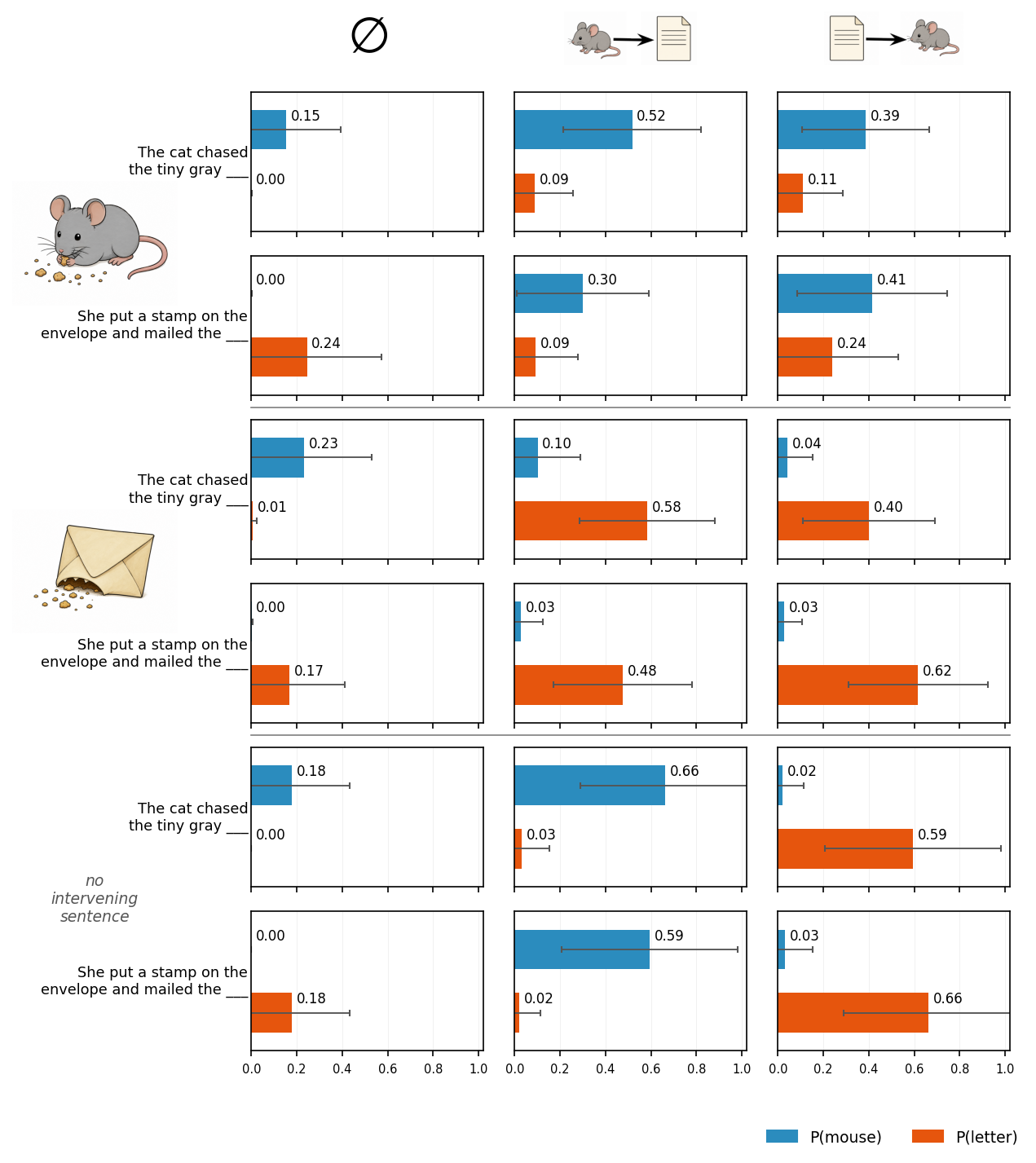}
  \caption{Qwen2.5-14B-Instruct: raw next-token probabilities. Rows: context
  sentence (icon) $\times$ query, plus two no-sentence rows; columns: redefinition
  condition (icon). Bars: P(mouse)$=w_1$, P(letter)$=w_2$; $\pm1$\,SD over pairs.}
  \label{fig:raw-14b}
\end{figure}

Figure~\ref{fig:raw-14b} gives the underlying raw
next-token probabilities, with no normalisation. The grid
adds, relative to the main figures, two no-sentence rows which are used as controls in the main text.

\newpage
\section{Prompt phrasing results}
\label{sec:phrasing-result}

\providecommand{\vdcell}[2]{\shortstack{#1\\[1pt]{\tiny$\pm$#2}}}

\begin{table}[h]
  \centering
  \small
  \setlength{\tabcolsep}{5pt}
  \begin{tabular}{cc cc cc cc}
    \toprule
    & & \multicolumn{2}{c}{$\varnothing$} & \multicolumn{2}{c}{\raisebox{-.35\height}{\includegraphics[height=5mm]{images//mouse_means_letter.jpeg}}} & \multicolumn{2}{c}{\raisebox{-.35\height}{\includegraphics[height=5mm]{images//letter_means_mouse.jpeg}}} \\
    \cmidrule(lr){3-4}\cmidrule(lr){5-6}\cmidrule(lr){7-8}
    context & query & \raisebox{-.35\height}{\includegraphics[height=3.5mm]{images//mouse.jpeg}} & \raisebox{-.35\height}{\includegraphics[height=3.5mm]{images//letter.jpeg}} & \raisebox{-.35\height}{\includegraphics[height=3.5mm]{images//mouse.jpeg}} & \raisebox{-.35\height}{\includegraphics[height=3.5mm]{images//letter.jpeg}} & \raisebox{-.35\height}{\includegraphics[height=3.5mm]{images//mouse.jpeg}} & \raisebox{-.35\height}{\includegraphics[height=3.5mm]{images//letter.jpeg}} \\
    \midrule
    \raisebox{-.35\height}{\includegraphics[height=5mm]{images//mouse_eats_crumbs.jpeg}} & \raisebox{-.35\height}{\includegraphics[height=3.5mm]{images//mouse.jpeg}} & \vdcell{.10}{.14} & \vdcell{.00}{.00} & \vdcell{.07}{.20} & \vdcell{.01}{.02} & \vdcell{.34}{.19} & \vdcell{-.28}{.26} \\
    \raisebox{-.35\height}{\includegraphics[height=5mm]{images//mouse_eats_crumbs.jpeg}} & \raisebox{-.35\height}{\includegraphics[height=3.5mm]{images//letter.jpeg}} & \vdcell{.00}{.01} & \vdcell{.07}{.10} & \vdcell{.05}{.22} & \vdcell{.02}{.02} & \vdcell{.30}{.23} & \vdcell{-.29}{.25} \\
    \raisebox{-.35\height}{\includegraphics[height=5mm]{images//letter_eats_crumbs.jpeg}} & \raisebox{-.35\height}{\includegraphics[height=3.5mm]{images//letter.jpeg}} & \vdcell{.00}{.00} & \vdcell{.10}{.19} & \vdcell{-.36}{.28} & \vdcell{.50}{.21} & \vdcell{.00}{.01} & \vdcell{.16}{.23} \\
    \raisebox{-.35\height}{\includegraphics[height=5mm]{images//letter_eats_crumbs.jpeg}} & \raisebox{-.35\height}{\includegraphics[height=3.5mm]{images//mouse.jpeg}} & \vdcell{.04}{.08} & \vdcell{.01}{.02} & \vdcell{-.44}{.28} & \vdcell{.52}{.27} & \vdcell{.00}{.02} & \vdcell{.15}{.28} \\
    \bottomrule
  \end{tabular}
  \caption{No-sentence-normalised sentence effect $\Delta P$ ($P_{\text{with sentence}}-P_{\text{no sentence}}$) for Qwen2.5-14B-Instruct under the \texttt{scope\_global} wording (8-word variant subset). Rows: context sentence $\times$ query word; columns: redefinition direction $\times$ queried word. Each cell: mean $\Delta P$ over the 56 pairs; $\pm$ line is the SD over pairs.}
  \label{tab:vardelta-scope-global}
\end{table}

\begin{table}[h]
  \centering
  \small
  \setlength{\tabcolsep}{5pt}
  \begin{tabular}{cc cc cc cc}
    \toprule
    & & \multicolumn{2}{c}{$\varnothing$} & \multicolumn{2}{c}{\raisebox{-.35\height}{\includegraphics[height=5mm]{images//mouse_means_letter.jpeg}}} & \multicolumn{2}{c}{\raisebox{-.35\height}{\includegraphics[height=5mm]{images//letter_means_mouse.jpeg}}} \\
    \cmidrule(lr){3-4}\cmidrule(lr){5-6}\cmidrule(lr){7-8}
    context & query & \raisebox{-.35\height}{\includegraphics[height=3.5mm]{images//mouse.jpeg}} & \raisebox{-.35\height}{\includegraphics[height=3.5mm]{images//letter.jpeg}} & \raisebox{-.35\height}{\includegraphics[height=3.5mm]{images//mouse.jpeg}} & \raisebox{-.35\height}{\includegraphics[height=3.5mm]{images//letter.jpeg}} & \raisebox{-.35\height}{\includegraphics[height=3.5mm]{images//mouse.jpeg}} & \raisebox{-.35\height}{\includegraphics[height=3.5mm]{images//letter.jpeg}} \\
    \midrule
    \raisebox{-.35\height}{\includegraphics[height=5mm]{images//mouse_eats_crumbs.jpeg}} & \raisebox{-.35\height}{\includegraphics[height=3.5mm]{images//mouse.jpeg}} & \vdcell{.10}{.14} & \vdcell{.00}{.00} & \vdcell{-.15}{.16} & \vdcell{.02}{.03} & \vdcell{.20}{.15} & \vdcell{-.21}{.21} \\
    \raisebox{-.35\height}{\includegraphics[height=5mm]{images//mouse_eats_crumbs.jpeg}} & \raisebox{-.35\height}{\includegraphics[height=3.5mm]{images//letter.jpeg}} & \vdcell{.00}{.01} & \vdcell{.07}{.10} & \vdcell{-.08}{.18} & \vdcell{.02}{.03} & \vdcell{.11}{.15} & \vdcell{-.19}{.22} \\
    \raisebox{-.35\height}{\includegraphics[height=5mm]{images//letter_eats_crumbs.jpeg}} & \raisebox{-.35\height}{\includegraphics[height=3.5mm]{images//letter.jpeg}} & \vdcell{.00}{.00} & \vdcell{.10}{.19} & \vdcell{-.30}{.23} & \vdcell{.29}{.18} & \vdcell{.00}{.01} & \vdcell{-.10}{.22} \\
    \raisebox{-.35\height}{\includegraphics[height=5mm]{images//letter_eats_crumbs.jpeg}} & \raisebox{-.35\height}{\includegraphics[height=3.5mm]{images//mouse.jpeg}} & \vdcell{.04}{.08} & \vdcell{.01}{.02} & \vdcell{-.35}{.21} & \vdcell{.21}{.17} & \vdcell{.02}{.02} & \vdcell{-.14}{.20} \\
    \bottomrule
  \end{tabular}
  \caption{Prompt-variation results for Qwen2.5-14B-Instruct (\texttt{scope\_sentence\_only} wording).}
  \label{tab:vardelta-scope-sentence-only}
\end{table}

\begin{table}[h]
  \centering
  \small
  \setlength{\tabcolsep}{5pt}
  \begin{tabular}{cc cc cc cc}
    \toprule
    & & \multicolumn{2}{c}{$\varnothing$} & \multicolumn{2}{c}{\raisebox{-.35\height}{\includegraphics[height=5mm]{images//mouse_means_letter.jpeg}}} & \multicolumn{2}{c}{\raisebox{-.35\height}{\includegraphics[height=5mm]{images//letter_means_mouse.jpeg}}} \\
    \cmidrule(lr){3-4}\cmidrule(lr){5-6}\cmidrule(lr){7-8}
    context & query & \raisebox{-.35\height}{\includegraphics[height=3.5mm]{images//mouse.jpeg}} & \raisebox{-.35\height}{\includegraphics[height=3.5mm]{images//letter.jpeg}} & \raisebox{-.35\height}{\includegraphics[height=3.5mm]{images//mouse.jpeg}} & \raisebox{-.35\height}{\includegraphics[height=3.5mm]{images//letter.jpeg}} & \raisebox{-.35\height}{\includegraphics[height=3.5mm]{images//mouse.jpeg}} & \raisebox{-.35\height}{\includegraphics[height=3.5mm]{images//letter.jpeg}} \\
    \midrule
    \raisebox{-.35\height}{\includegraphics[height=5mm]{images//mouse_eats_crumbs.jpeg}} & \raisebox{-.35\height}{\includegraphics[height=3.5mm]{images//mouse.jpeg}} & \vdcell{.10}{.14} & \vdcell{.00}{.00} & \vdcell{-.02}{.24} & \vdcell{-.09}{.18} & \vdcell{.22}{.23} & \vdcell{-.13}{.18} \\
    \raisebox{-.35\height}{\includegraphics[height=5mm]{images//mouse_eats_crumbs.jpeg}} & \raisebox{-.35\height}{\includegraphics[height=3.5mm]{images//letter.jpeg}} & \vdcell{.00}{.01} & \vdcell{.07}{.10} & \vdcell{-.04}{.13} & \vdcell{-.01}{.21} & \vdcell{.25}{.16} & \vdcell{-.20}{.23} \\
    \raisebox{-.35\height}{\includegraphics[height=5mm]{images//letter_eats_crumbs.jpeg}} & \raisebox{-.35\height}{\includegraphics[height=3.5mm]{images//letter.jpeg}} & \vdcell{.00}{.00} & \vdcell{.10}{.19} & \vdcell{-.15}{.19} & \vdcell{.24}{.26} & \vdcell{-.13}{.17} & \vdcell{.11}{.27} \\
    \raisebox{-.35\height}{\includegraphics[height=5mm]{images//letter_eats_crumbs.jpeg}} & \raisebox{-.35\height}{\includegraphics[height=3.5mm]{images//mouse.jpeg}} & \vdcell{.04}{.08} & \vdcell{.01}{.02} & \vdcell{-.24}{.24} & \vdcell{.26}{.17} & \vdcell{-.07}{.21} & \vdcell{.05}{.16} \\
    \bottomrule
  \end{tabular}
  \caption{Prompt-variation results for Qwen2.5-14B-Instruct (\texttt{denotational} wording).}
  \label{tab:vardelta-denotational}
\end{table}

\begin{table}[h]
  \centering
  \small
  \setlength{\tabcolsep}{5pt}
  \begin{tabular}{cc cc cc cc}
    \toprule
    & & \multicolumn{2}{c}{$\varnothing$} & \multicolumn{2}{c}{\raisebox{-.35\height}{\includegraphics[height=5mm]{images//mouse_means_letter.jpeg}}} & \multicolumn{2}{c}{\raisebox{-.35\height}{\includegraphics[height=5mm]{images//letter_means_mouse.jpeg}}} \\
    \cmidrule(lr){3-4}\cmidrule(lr){5-6}\cmidrule(lr){7-8}
    context & query & \raisebox{-.35\height}{\includegraphics[height=3.5mm]{images//mouse.jpeg}} & \raisebox{-.35\height}{\includegraphics[height=3.5mm]{images//letter.jpeg}} & \raisebox{-.35\height}{\includegraphics[height=3.5mm]{images//mouse.jpeg}} & \raisebox{-.35\height}{\includegraphics[height=3.5mm]{images//letter.jpeg}} & \raisebox{-.35\height}{\includegraphics[height=3.5mm]{images//mouse.jpeg}} & \raisebox{-.35\height}{\includegraphics[height=3.5mm]{images//letter.jpeg}} \\
    \midrule
    \raisebox{-.35\height}{\includegraphics[height=5mm]{images//mouse_eats_crumbs.jpeg}} & \raisebox{-.35\height}{\includegraphics[height=3.5mm]{images//mouse.jpeg}} & \vdcell{.10}{.14} & \vdcell{.00}{.00} & \vdcell{.01}{.24} & \vdcell{.00}{.01} & \vdcell{.12}{.10} & \vdcell{-.17}{.16} \\
    \raisebox{-.35\height}{\includegraphics[height=5mm]{images//mouse_eats_crumbs.jpeg}} & \raisebox{-.35\height}{\includegraphics[height=3.5mm]{images//letter.jpeg}} & \vdcell{.00}{.01} & \vdcell{.07}{.10} & \vdcell{.01}{.23} & \vdcell{.00}{.01} & \vdcell{.13}{.12} & \vdcell{-.12}{.22} \\
    \raisebox{-.35\height}{\includegraphics[height=5mm]{images//letter_eats_crumbs.jpeg}} & \raisebox{-.35\height}{\includegraphics[height=3.5mm]{images//letter.jpeg}} & \vdcell{.00}{.00} & \vdcell{.10}{.19} & \vdcell{-.33}{.24} & \vdcell{.36}{.16} & \vdcell{.00}{.01} & \vdcell{.08}{.29} \\
    \raisebox{-.35\height}{\includegraphics[height=5mm]{images//letter_eats_crumbs.jpeg}} & \raisebox{-.35\height}{\includegraphics[height=3.5mm]{images//mouse.jpeg}} & \vdcell{.04}{.08} & \vdcell{.01}{.02} & \vdcell{-.38}{.19} & \vdcell{.35}{.18} & \vdcell{.00}{.01} & \vdcell{.00}{.26} \\
    \bottomrule
  \end{tabular}
  \caption{Prompt-variation results for Qwen2.5-14B-Instruct (\texttt{comprehension\_only} wording).}
  \label{tab:vardelta-comprehension-only}
\end{table}

\end{document}